# Development and evaluation of a 3D model observer with nonlinear spatiotemporal contrast sensitivity

Ali R. N. Avanaki[a], Kathryn S. Espig[a], Andrew D. A. Maidment[c],
Cédric Marchessoux[b], Predrag R. Bakic[c], Tom R. L. Kimpe[b]

[a]Barco Healthcare, Beaverton, OR; [b]Barco Healthcare, Kortrijk, Belgium;
[c]University of Pennsylvania, Philadelphia, PA

**ABSTRACT**

We investigate improvements to our 3D model observer with the goal of better matching human observer performance as a function of viewing distance, effective contrast, maximum luminance, and browsing speed. Two nonlinear methods of applying the human contrast sensitivity function (CSF) to a 3D model observer are proposed, namely the Probability Map (PM) and Monte Carlo (MC) methods. In the PM method, the visibility probability for each frequency component of the image stack, $p$, is calculated taking into account Barten's spatiotemporal CSF, the component modulation, and the human psychometric function. The probability $p$ is considered to be equal to the perceived amplitude of the frequency component and thus can be used by a traditional model observer (e.g., LG-msCHO) in the space-time domain. In the MC method, each component is randomly kept with probability $p$ or discarded with 1-$p$. The amplitude of the retained components is normalized to unity. The methods were tested using DBT stacks of an anthropomorphic breast phantom processed in a comprehensive simulation pipeline. Our experiments indicate that both the PM and MC methods yield results that match human observer performance better than the linear filtering method as a function of viewing distance, effective contrast, maximum luminance, and browsing speed.

**Keywords:** Human visual system properties, spatiotemporal contrast sensitivity function, psychometric function, channelized Hotelling observer, and anthropomorphic numerical observer, virtual clinical trials.

## 1. INTRODUCTION

The human visual system (HVS) does not respond equally to all spatial, temporal, or spatiotemporal stimuli. Specifically, the just-noticeable modulation (a.k.a. visibility threshold) varies for contrast excitations at different spatiotemporal frequencies. The inverse of the contrast visibility threshold is called the spatiotemporal contrast sensitivity function (stCSF) [1, 3]; this function is used to model the main properties of the HVS.

Our goal is to enhance a frequently used 3D observer model by making it perform more like a human observer in a variety of aspects. A good candidate for an anthropomorphic observer model, we believe, should behave like a human observer with regard to changes in viewing conditions, including viewing distance, browsing speed, contrast and luminance. Our approach to the problem of designing an anthropomorphic observer model involves modeling human perception by simulating the HVS properties.

In the past [7], we modeled the HVS as the modulation of each spatiotemporal frequency component of the input normalized by the just-noticeable modulation at that spatiotemporal frequency. We applied this observer model to the task of viewing a digital breast tomosynthesis (DBT) stack browsed in time. The underlying assumption therein was that each spatiotemporal frequency component of an image could be processed independently (i.e., no contrast masking). Specifically, we calculated the ratio $m/(S^{-1})$, where $m$ is the modulation of the spatiotemporal frequency component, defined as the component's amplitude normalized by the average input luminance, and $S^{-1}$ is the visibility threshold predicted by Barten's model [1] at that spatiotemporal frequency. The ratio above can be also written as $mS$, showing that this processing is equivalent to linear filtering (Section 2.2). This approach predicts that the detection performance will peak (i.e., reach a maximum) at a specific browsing speed, as expected from human observer studies [7]. However, linear filtering cannot explain why humans have improved detection performance with increased contrast or why performance peaks at a specific viewing distance. Thus, it is necessary to consider non-linear methods.

We hypothesized that the inherent non-linearity of the HVS, given by the psychometric function (Fig. 1), needed to be modeled. One way of modeling the psychometric function is by inclusion of an additive noise source in the observer pipeline [2, 8]. This common method reduces the influence of weak signals, while preserving the influence of strong signals. A weakness of this approach is the requirement of an *ad hoc* calibration procedure, used to determine the spectrum of noise required to match the model's behavior to that of the HVS. Typically a white spectrum is used, but this is not well justified.

Significant predicate work exists in the design of non-linear anthropomorphic model observers. Park *et al.* integrated Barten's spatial CSF (Ch. 3 of [1]) into a channelized model observer [2]. This work differs from ours because the perceived (i.e., contrast-thresholded) images are approximated to a single spatial frequency (Eq. 18 of [2]). We believe that this is unlikely to predict the behavior of a human observer when viewing images consisting of a multitude of spatiotemporal frequencies, typical of the signals considered in this paper.

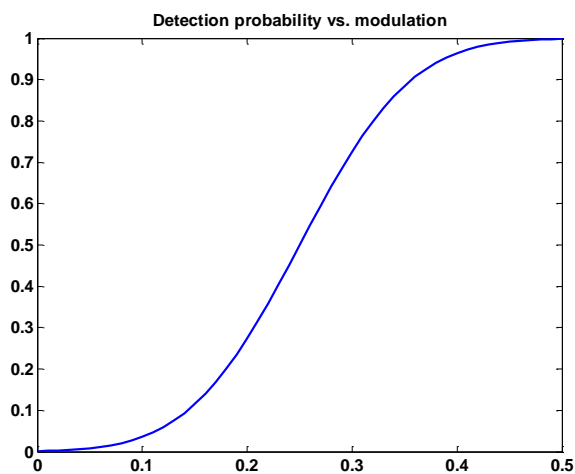

**Fig. 1.** Detection probability depicted against modulation, for a visibility threshold of 0.25.

In [8], Zhang *et al.* studied the effect on detection performance of having a non-linearity in a model observer pipeline. They state that "the channelized Hotelling (observer) with a number of nonlinear components was a slightly better predictor of human performance than the addition of internal noise." However, they conclude that the small impact of having a non-linearity model in the pipeline on the results does not justify the additional computational costs. It is worth noting that the non-linearity model (Watson's excitatory, and excitatory/inhibitory models [13]), the modality (2D x-rays), and the task (comparison of JPEG and JPEG2000) in [8] were different from those of this work.

In this paper, we examine two non-linear ways of affecting the stCSF. In the first way, we use the psychometric function (Fig. 1 and Eq. 10) to calculate the probability of visibility for a spatiotemporal frequency component and use that probability as the perceived amplitude for that component (Section 2.3). We also consider a pseudo-random approach (Section 2.4), where we keep spatiotemporal frequency components that are deemed visible with the abovementioned visibility probability and discard the remaining components (i.e., to force their amplitude to zero).

## 2. METHODS

### 2.1 Simulation platform

Validation of any imaging system is challenging due to the large number of system parameters that must be considered. Conventional methods involving clinical trials are limited by cost and duration, and in the instance of systems using ionizing radiation, the requirement for the repeated irradiation of volunteers. We are strong proponents of a preclinical alternative, in the form of Virtual Clinical Trials (VCTs), which are based upon models of human anatomy, image acquisition, display and processing, and image analysis and interpretation.

In this work, synthetic breast images were generated using the breast anatomy and imaging simulation pipeline developed at the University of Pennsylvania (UPenn). Fig. 2 illustrates the structure of this pipeline. Normal breast anatomy is simulated based upon a recursive partitioning algorithm using octrees [6]. Calcified breast lesions may be included in the software breast phantom, by automatic insertion of microcalcification clusters [14]. Phantom deformation due to clinical breast positioning and compression is simulated using a finite element model [15]. DBT image acquisition is simulated by ray tracing projections through the phantoms, assuming a polyenergetic x-ray beam without scatter, and an ideal detector model. Reconstructed breast images are obtained using the Real-Time Tomography image reconstruction and processing method [16]. The simulation modules in the pipeline are interconnected using an XML-based dynamic parsimonious data representation, providing a high level of control and flexibility of the simulation.

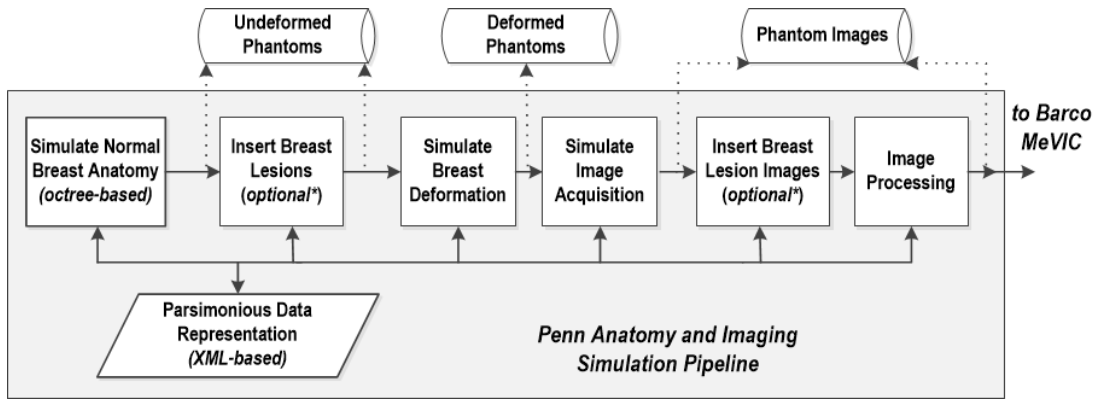

**Fig. 2.** Flow chart of the pipeline for breast anatomy and imaging simulation.

The display and virtual observer simulation is implemented in MEVIC (Medical Virtual Imaging Chain) [4], an extensible C++ platform developed for medical image processing and visualization at Barco. The structure of this pipeline is illustrated in Fig. 3. DBT stack datasets (volumes of interest) with and without simulated lesions, generated using the UPenn pipeline, are input to the display and virtual observer simulation pipeline. For the experiments reported here, 3296 reconstructed 64x64x32 DBT image stacks are used, half with lesions and half without. Slices from a sample stack are shown in Fig. 4. Each stack is first decomposed into its spatiotemporal frequency components using a 3D fast Fourier transform (FFT). The frequency components are processed independently; thus one component does not affect the visibility of another (i.e., no contrast masking). The methods proposed in this paper (Sections 2.2 - 2.4) are implemented in the dotted block in Fig. 3 to determine the perceived amplitude of each frequency component. Then, an inverse 3D FFT is applied to the perceived amplitudes to transform the perceived stack into the space-time domain. Finally, the results are input to a multi-slice channelized Hotelling observer (msCHO) developed by Platiša *et al* [5]. Further details of the simulation are provided in our previous work [7].

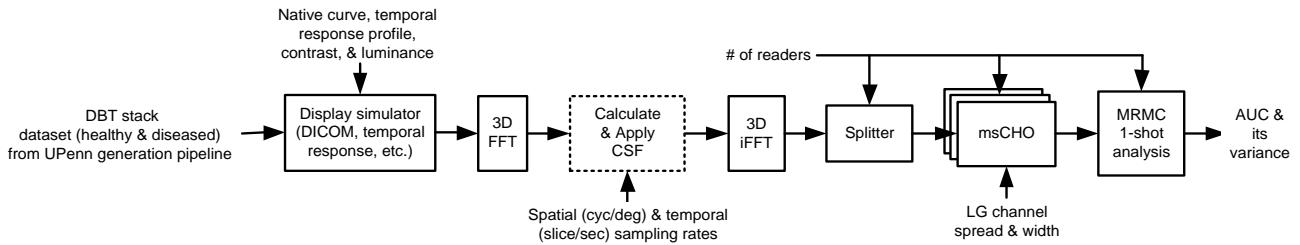

**Fig. 3.** Block diagram of the display and virtual observer simulation. The methods proposed in this paper are used in the dotted block.

## 2.2 Linear filtering

We implemented linear filtering (LF) in much the same way as our previous work [7]. The amplitude of the signal component with a spatiotemporal frequency of ($u$, $w$) is multiplied by the stCSF at that frequency, [$S(u, w)$]. Following the work of Barten, Eq. 5.2 in [1] is given by

$$S(u,w) = \frac{M_{opt}(u)}{k\sqrt{\frac{2}{T}\left(\frac{1}{X_0^2}+\frac{1}{X_{max}^2}+\frac{u^2}{N_{max}^2}\right)\left(\frac{1}{\eta p E}+\frac{\Phi_0}{[H_1(w)\{1-H_2(w)F(u)\}]^2}\right)}} \quad (1)$$

where $u$ and $w$ are the spatial (in cycle/deg) and temporal (in cycle/sec) frequencies, respectively. For completeness, all terms of Eq. 1 are given below. From Eq. 3.21 in [1], we have

$$F(u) = 1 - \sqrt{1-\exp(-(u/u_0)^2)} \quad (2)$$

with $u_0 = 7$. For $H_i(w)$, $i = 1, 2$, we use Eq. 5.5 in [1]:

$$H_i(w) = \sqrt{\left(1+(2\pi\tau_i w)^2\right)^{-n_i}}, \quad i=1,\,2 \quad (3)$$

with parameters given by Eq. 5.11 and Eq. 5.12 in [1]. Namely,

$$\tau_1 = \frac{\tau_{10}}{1+0.55\ln\left\{1+(1+D)^{0.6}\frac{E}{3.5}\right\}}, \quad (4)$$

$$\tau_2 = \frac{\tau_{20}}{1+0.37\ln\left\{1+\left(1+\frac{D}{3.2}\right)^5\frac{E}{120}\right\}} \quad (5)$$

where $n_1 = 7, n_2 = 4, \tau_{10} = 32e-3, \tau_{20} = 18e-3, D = 2X_0/\sqrt{\pi}$ are set per recommendations in [1]. Also,

$$M_{opt}(u) = \exp(-2(\pi\sigma u)^2), \quad (6)$$

$$\sigma = \frac{1}{60}\sqrt{\sigma_0^2 + (C_{ab}d)^2}, \quad (7)$$

$$E = \frac{\pi d^2 L}{4}\left(1-(d/9.7)^2+(d/12.4)^4\right), \text{ and} \quad (8)$$

$$d = 5 - 3\tanh\{0.4\ln(LX_0^2/40^2)\} \quad . \quad (9)$$

Here, $L$ is the average luminance of the object (over space & time, in cd/m$^2$), $X_0$ is the apparent size of the image (in degrees; dependent upon the image size in pixels, the pixel pitch, and the viewing distance, assuming a viewing axis orthogonal to the image plane), $d$ is the pupil diameter in mm, and $E$ is retinal illuminance, in Trolands. $M_{opt}$ is the optical modulation transfer function of the eye, and $\sigma$ is the standard deviation of the line-spread function, with $C_{ab} = 0.08$ and $\sigma_0 = 0.5$. The rest of parameters are: $\eta = 0.03$, $\Phi_0 = 3e-8$, $X_{max} = 12$, $N_{max} = 15$, $T = 0.1$, $p = 1.285e6$ (from Table 3.2 in [1]) and $k = 3$.

Our data includes two orthogonal spatial frequencies, $u_1$ and $u_2$. To calculate $u$, the single spatial frequency required for $S$ given by Eq. 1, we use $u = \sqrt{u_1^2 + u_2^2}$. As compared to [7], a 2x speed-up in computation speed was achieved by considering the fact that almost half of the data are complex conjugates of the other half as these input data are always real. This acceleration holds for the methods below as well. The other important difference with [7] is that each stack is modified using a linear mapping, so that it has at least one pixel at the maximum display luminance ($L_{max}$) and at least one pixel at the lowest display luminance ($L_{min} = L_{max}$ / contrast). Thus, each stack is displayed at the same effective contrast.

### 2.3 Probability map

For our second method, we developed a probability map in the spatiotemporal frequency domain. Starting with Eq. 2.2 of [1], it is possible to write the detection probability, $p$, for a spatiotemporal frequency component as

$$p = \frac{1}{2} + \frac{1}{2}\text{erf}\left(\frac{z}{\sqrt{2}}\right) \tag{10}$$

where erf( ) is the error function [11], and

$$z = k(mS - 1) \quad . \tag{11}$$

Here $k$ is the Crozier coefficient, and $m$ is the modulation at the frequency under consideration. $S$ is the stCSF given by Eq. 1 in Section 2.2. Eq. 11 differs slightly from Eq. 2.5 of [1] in terms of notation. The perceived amplitude of the component is then considered to be $p$, as given by Eq. 10.

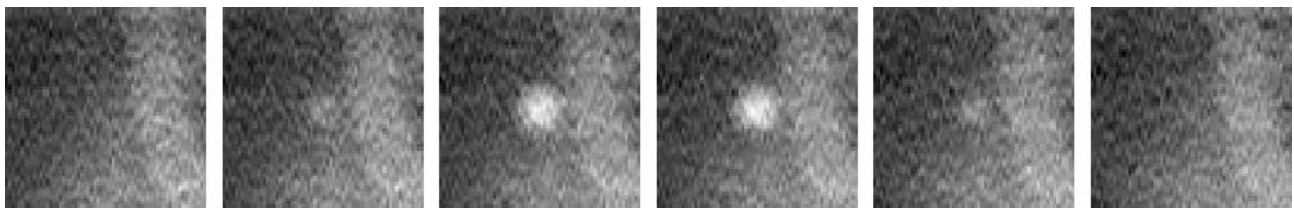

**Fig. 4**. Six central slices of a sample stack used in simulation. The insertion contrast for this image is increased to make the lesion visible for the purposes of publication.

Following Eq. 2.30 in [1], $m$ is calculated by normalizing the amplitude of the various frequency components by the average luminance. Because the normalized data are real-valued, we take advantage of the inherent symmetry of the discrete Fourier transform to reduce the number of components by approximately one-half. Note that this is different from our prior work in [7] and [10]. This is of particular importance in considering the MC method (see Section 2.4) because using our prior method conjugate components would be calculated separately, and thus it would be possible that the conjugate components would differ. This would result in the signal having a non-zero imaginary part after the inverse Fourier transform.

### 2.4 Monte Carlo

In our final approach, we use Monte Carlo (MC) methods to calculate each component of the perceived signal in the frequency domain. Each component of the signal is randomly kept with probability $p$ (as calculated in Section 2.3), or is set to zero with probability 1-$p$. Amplitudes of all of the retained components are normalized to unity since we do not want to differentiate between any of the components that are deemed visible.

Fig. 5 offers a graphical comparison of the LF, PM, and MC methods. Here, for easier visualization, we show a one-dimensional frequency rather than the 3D spatiotemporal frequency. A signal with four individual frequency components is shown for further simplicity. The signal (labeled $m$ for modulation) is given by the component amplitude divided by average luminance of the signal.

For the LF method (Section 2.2), the amplitude of each component is proportional to $mS$, where $S$ is given by Eq. 1. The value of $mS$ can be interpreted as being the multiple by which $m$ is larger than the visibility threshold ($S^{-1}$). For the PM method (Section 2.3), the magnitude of each component is proportional to the probability of visibility $p$ given by Eq. 10. The outcome of the MC (Section 2.4) method is not deterministic. For the 4-component example signal shown here, there are 16 ($=2^4$) possible outcomes. Three possible outcomes are shown. Since $p$ is largest for the first component and smaller for the other components, the most likely output (the first one shown) will include this first component and not the rest. The amplitude of the components, if kept, is set to unity.

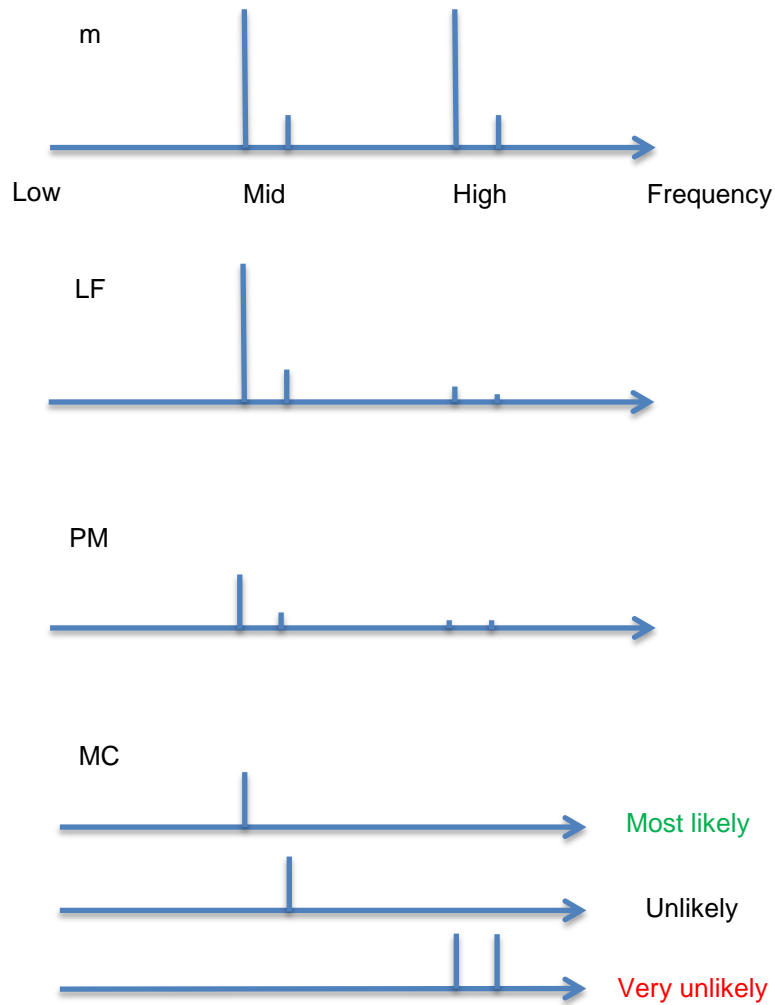

**Fig. 5**. A graphical example explaining the methods of Section 2. The frequency is shown as a single dimension for demonstration purposes. A sample signal consisting of four frequency components at mid- to high-frequencies is shown at the top; here, the height of each bar (each representing a single frequency component) is proportional to the modulation $m$ of the component. After modulation, the results of applying stCSF using the linear (LF) and nonlinear (PM, MC) methods are shown. Only three of the 16 possible outcomes of MC application are shown in decreasing order of likelihood for demonstration purposes.

# 3. RESULTS AND DISCUSSION

We have compared each method of Section 2 in response to various alterations to the display conditions: maximum display luminance, display contrast, spatial sampling frequency and browsing speed. Our goal is to identify which method(s) follow the expected human observer behavior for the various conditions.

AUC values and error bars (twice the standard deviation to give 95% confidence intervals) are calculated using the one-shot multi-reader multi-case (MRMC) method developed by Gallas [9] on the output of four virtual observers. The virtual observers are instances of the type 'b' msCHO (the middle diagram of Fig. 6 in [5]), in which the template calculated for the central slice is applied to all slices of interest and the results are used by a Hotelling observer with 15 LG channels and a spread of 10. For further details please see [7].

## 3.1 Luminance and contrast:

The results of detection performance as a function of the effective contrast are summarized in Fig. 6. To compare the methods, we express the results in terms of d' (detectability index) which is derived from the AUC as follows (Eq. 9 in [17])

$$d' = 2\,\text{erf}^{-1}(2\,\text{AUC} - 1) \qquad (12)$$

where $\text{erf}^{-1}(\ )$ is the inverse error function [11]. Other simulation settings include a browsing speed of 25 slice/sec, $L_{max}$=300 cd/m$^2$ (fixed for the variable contrast experiment), an effective contrast of C = 200 (fixed for the variable maximum luminance experiment), a spatial sampling rate of 7 pixels/degree, and large lesions (masses).

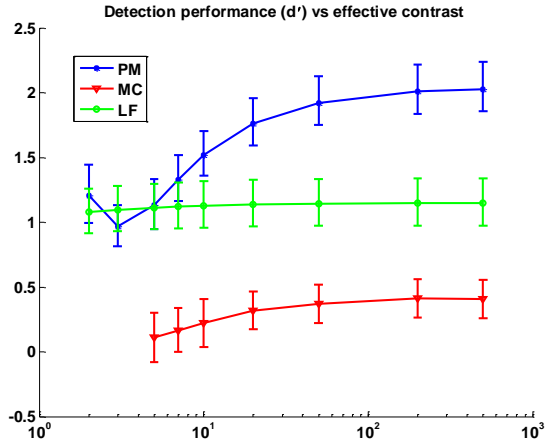

**Fig. 6.** Detection performance in d' depicted as a function of the effective contrast.

It is to be expected that the observer detection performance will improve with increased effective contrast. This observation trend with contrast can be seen in Fig. 6 for the PM and MC methods. However, consistent with our prior results [7], no trend with contrast is seen for the LF method.

The expected trend with maximum display luminance is less obvious, because maximum luminance is typically linked to display contrast. Thus as $L_{max}$ increases, the number of available perceptually distinct gray levels (JNDs) increases (from ~300 to ~700 for the luminance range used in our simulation). Our methods however fail to show this; detection performance remains constant using the MC method and decreases with increasing $L_{max}$ using PM and LF.

To explore this effect, we present the following example. Consider a two-frequency luminance excitation $f$ given by

$$f(x,y,t) = \frac{2+\cos(u_1 x + u_2 y + wt) + \cos(u'_1 x + u'_2 y + w't)}{4}\left(L_{max} - \frac{L_{max}}{C}\right) + \frac{L_{max}}{C} \quad , \qquad (13)$$

which oscillates between $L_{max}$ and $L_{min} = L_{max}/C$ with spatiotemporal frequencies $(u_1, u_2, w)$ and $(u'_1, u'_2, w')$. $C$ is the effective contrast. The average luminance, $L$, is

$$L = \frac{1}{2}\left(L_{max} - \frac{L_{max}}{C}\right) + \frac{L_{max}}{C} \quad (14)$$

The modulation is given by the amplitude of variations, $\frac{1}{4}\left(L_{max} - \frac{L_{max}}{C}\right)$, for both cosine components, divided by $L$. Therefore, we have

$$m(u_1, u_2, w) = m(u'_1, u'_2, w') = \frac{\frac{1}{4}\left(L_{max} - \frac{L_{max}}{C}\right)}{\frac{1}{2}\left(L_{max} - \frac{L_{max}}{C}\right) + \frac{L_{max}}{C}} = \frac{1-C^{-1}}{2(1+C^{-1})} \quad (15)$$

Note that the modulation ends up being a function of effective contrast only. This, however, does not mean that detecting an excitation is independent of $L_{max}$. In practice, $L_{min}$, is a function of display black leakage and viewing conditions (veiling glare and/or surround light, etc.), and is commonly considered constant. Hence, increasing $L_{max}$ should increase the effective contrast directly. Moreover, Barten's CSF which determines whether a given modulation is visible, is a function of $L$ which depends on $L_{max}$. As a result, $L_{max}$ affects $S$ via $L$ (the average luminance), and the CSF changes with $L$ as a function of spatial and temporal frequencies, and target size, which is a function of viewing distance in this scenario. Hence, one can expect that, at fixed contrast, the detection performance increases with $L_{max}$. This is seen in Fig. 7, where the CSF is shown as a function of $L$ ; in which case, the high-$L$ CSF is always larger than low-$L$ CSF.

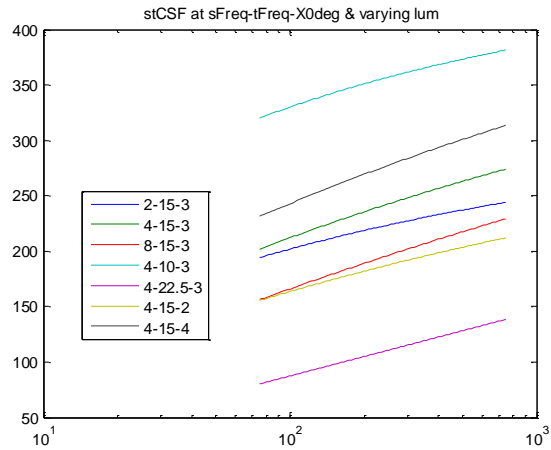

**Fig. 7**. When varying the average target luminance $L$ , the CSF increase depends on other parameters (spatiotemporal frequency) and target size.

### 3.2 Spatial sampling frequency (SSR):

Spatial sampling frequency, in pixels per degree, depends on the display's pixel pitch and the viewing distance, assuming orthogonal viewing. For a given display resolution (i.e., fixed pixel pitch), SSR becomes a function of viewing distance alone. Assuming that 64×64 slices are viewed in a 9 cm² (3 cm x 3 cm) area on the display, the viewing distance is given by

$$d = \frac{3\ cm}{2\tan\left(\frac{64\ \pi}{360\ SSR}\right)} \quad (16)$$

The results of detection performance as a function of viewing distance are summarized in Fig. 8. Other simulation settings are: a browsing speed of 25 slice/sec, $L_{max}$=300 cd/m², $C = 200$, and large lesions (masses).

Based on our daily visual experience, it is clear that too small or too large a viewing distance is not optimum for distinguishing the details of a displayed visual signal. That is because at an overly short distance, information bearing pixels are moved out of the foveal visual, and sub-pixel-structure (the areas between luminance elements) introduce visual noise. When viewed at too great a distance, a signal's information content is observed at spatial-frequencies that are excessively high and thus are suppressed by the stCSF. By similar reasoning, it is also well known that the optimum viewing distance is a function of the detection target (lesion) [12].

As can be seen in Fig. 8, both the PM and MC methods show the expected trend with viewing distance, demonstrating reduced performance at short and long view distances. By contrast, the LF method does not produce results consistent with human observer performance.

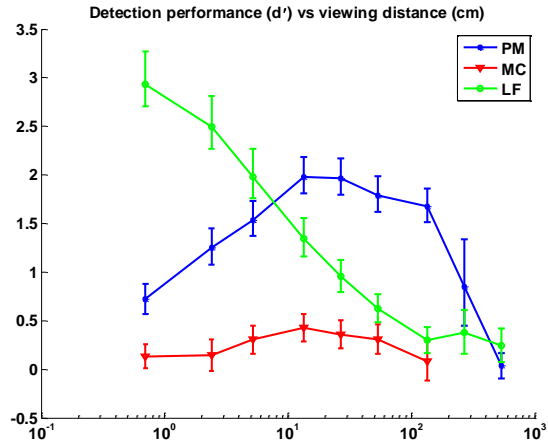

**Fig. 8.** Detection performance in d' as a function of viewing distance.

### 3.3 Browsing speed:

The results of detection performance as a function of browsing speed are summarized in Fig. 9. Other simulation settings are: SSR=7 pixel/deg, $L_{max}$=300 cd/m$^2$, C = 200, and large lesions (masses). Based on daily visual experience, one knows that a browsing speed that is too slow or too fast is not optimum for distinguishing details of a displayed spatiotemporal visual signal. That is because the temporal frequency content of the stack is determined by the browsing speed. When the stack is browsed too slowly (or too quickly), the temporal frequency content is shifted towards lower (or higher) frequencies, where the stCSF is lower. All three methods show a reduction in performance at low and high browsing speed, as expected.

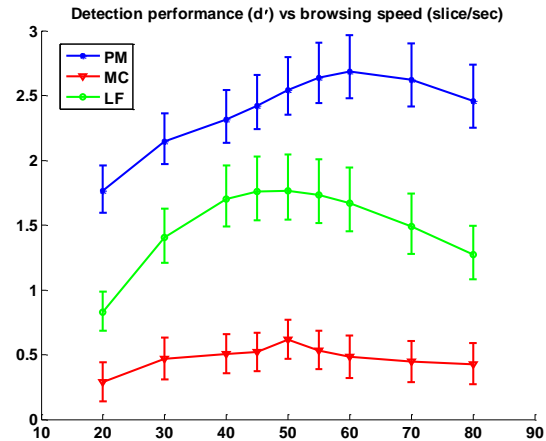

**Fig. 9.** Detection performance in d' depicted as a function of browsing speed.

### 3.4 Summary

The results presented in Fig. 6, 8, and 9 are shown together in Fig. 10. In this graph, we present the detection performance normalized in terms of d'/d'$_{max}$, in order to be able to better compare the trends of observer performance with the various factors examined. These results are also summarized in Table 1, where the expected performance of the human observer is compared to the measured performance of the various machine observers.

As can be seen, both the PM and MC observers match human performance better than the LF method. The main difference between the PM and MC observers is the absolute difference between the d' values; the d'/d'$_{max}$ values match much more closely. This difference requires that an internal noise source be added to the PM method to match human observer performance in absolute terms, while the MC method should avoid the need for a noise source, if the psychometric function can be controlled to match human performance. Intrinsically, it is appealing to have a method that avoids the introduction of an *ad hoc* noise source. However, we have yet to show that human performance can be matched exactly. This is the subject of future work.

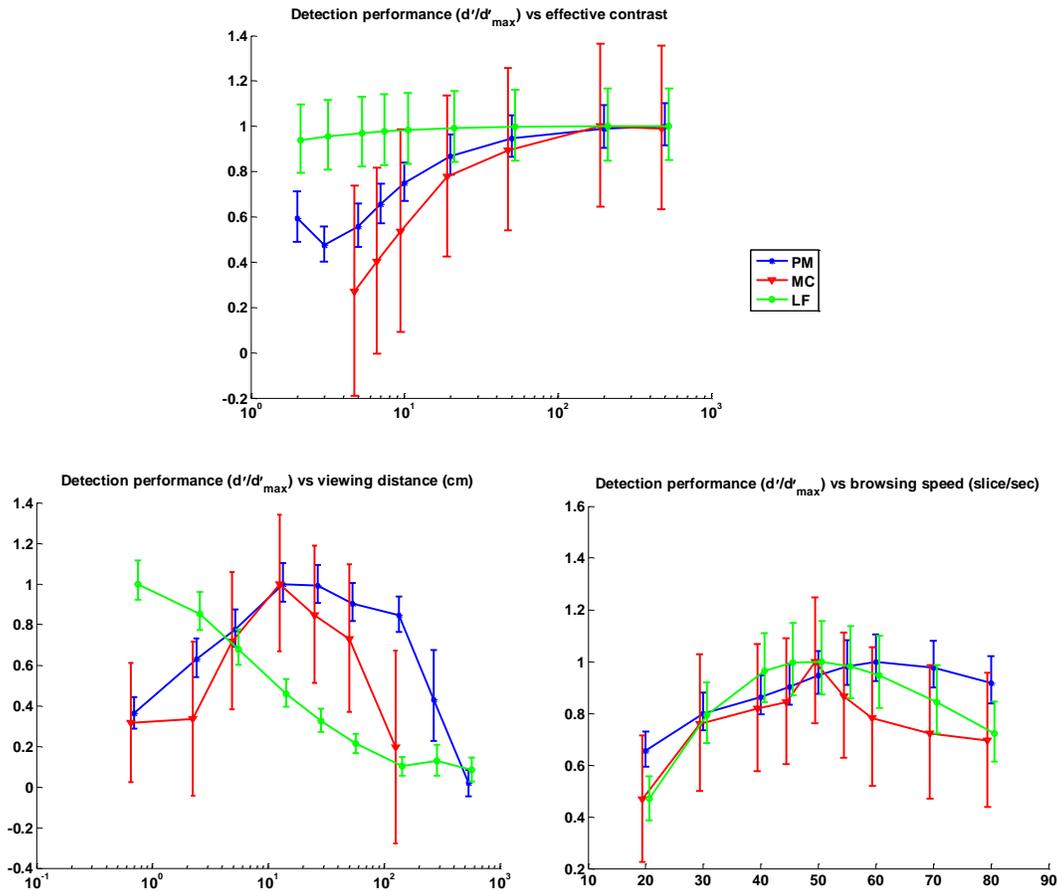

**Fig. 10.** Detection performance in terms of d'/d'$_{max}$ depicted as a function, effective contrast (top), viewing distance (bottom left) and browsing speed (bottom right).

**Table 1.**  A summary of the machine observer results is presented.  The detection performance of the proposed methods is compared to the behavior expected from a human observer.  Entries in ***italics*** match human observers.

| Trend | Human | LF | PM | MC |
|---|---|---|---|---|
| Effective contrast | *increasing* | constant | *increasing* | *increasing* |
| $L_{max}$ (at fixed contrast) | *increasing* | decreasing | decreasing | constant |
| Viewing distance | *peaked* | decreasing | *peaked* | *peaked* |
| Browsing speed | *peaked* | *peaked* | *peaked* | *peaked* |

## 4. CONCLUSION

Our results indicate that integration of Barten's stCSF with a model observer in a nonlinear fashion, including the PM (Section 2.2) and MC (Section 2.3) approaches, yield a machine observer that better conforms to the detection performance expected from human observers with respect to browsing speed, viewing distance, effective contrast and luminance.

This project remains a work in progress.  The following are planned for the near future: (1) investigate why our model does not correctly predict detection performance change with maximum display luminance, and alter the model accordingly; (2) perform psychophysical experiments to validate the results of our work in this paper and to select the CSF method which best matches human observers; and (3) include a contrast masking model to mimic human observation trends with respect to background complexity.

## ACKNOWLEDGEMENT

This work is supported by the US National Institutes of Health (grant 1R01CA154444).  Ali Avanaki would like to thank Albert Xthona, Kyle Myers, Miguel Eckstein, Bastian Piepers and Ljiljana Platiša.  Dr. Maidment is on the Scientific Advisory Board of Real-Time Tomography, LLC.

## REFERENCES

[1] P.G.J. Barten, *Contrast sensitivity of the human eye and its effects on image quality*, SPIE Optical Engineering Press, Bellingham, WA, 1999.
[2] S. Park, A. Badano, B.D. Gallas, K. J. Myers, "Incorporating Human Contrast Sensitivity in Model Observers for Detection Tasks," *IEEE Trans. Medical Imaging*, vol. 28, pp. 339-347, 2009.
[3] S. Daly, "Engineering observations from spatiovelocity and spatiotemporal visual models," *Proc. of SPIE*, vol. 3299, pp. 180-191, 1998.
[4] C. Marchessoux, T. R. L. Kimpe, and T. Bert, "A virtual image chain for perceived and clinical image quality of medical display," *J. of Display Technology*, vol. 4, pp. 356–368, 2008.
[5] L. Platiša, B. Goossens, E. Vansteenkiste, S. Park, B. Gallas, A. Badano and W. Philips, "Channelized hotelling observers for the assessment of volumetric imaging data sets," *J. of Optical Society of America A*, vol. 28, pp. 1145 – 1163, 2011.
[6] D. Pokrajac, A.D.A. Maidment, P.R. Bakic, "Optimized Generation of High Resolution Breast Anthropomorphic Software Phantoms," *Medical Physics*, vol. 39, pp. 2290-2302, April 2012.
[7] A.N. Avanaki, K.S. Espig, C. Marchessoux, E.A. Krupinski, P.R. Bakic, T.R.L. Kimpe, A.D.A. Maidment, "Integration of spatio-temporal contrast sensitivity with a multi-slice channelized Hotelling observer," *Proc. SPIE Medical Imaging*, 2013.
[8] Y. Zhang, B.T. Pham, M.P. Eckstein, "The Effect of Nonlinear Human Visual System Components on Performance of a Channelized Hotelling Observer Model in Structured Backgrounds," *IEEE Trans. on Medical Imaging*, vol. 25, pp. 1348-1362, 2006.
[9] B.D. Gallas, "One-shot estimate of MRMC variance: AUC," *Acad Radiol.*, vol. 13, pp. 353-362, 2006.
[10] A.N. Avanaki, K.S. Espig, C. Marchessoux, E.A. Krupinski, P.R. Bakic, T.R.L. Kimpe, A.D.A. Maidment, "On modeling the effects of display contrast and luminance in a spatio-temporal numerical observer," *MIPS* presentation, Washington DC, 2013.

[11] http://en.wikipedia.org/wiki/Error_function, accessed February 2014.
[12] C. Marchessoux, based on a clinical study with new products, private communication, August 2013.
[13] A.B. Watson and J.A. Solomon, "Model of visual contrast gain control and pattern masking," *J. Opt. Soc. Amer. A*, vol. 14, pp. 2379–2391, September 1997.
[14] V. Shankla, D. Pokrajac, S. Weinstein, M. De Leo, C. Tuite, R. Roth, E. Conant, A.D.A. Maidment, P.R. Bakic, "Automatic Insertion of Simulated Microcalcification Clusters in a Software Breast Phantom," *Proc. of SPIE,* Medical Imaging 2014.
[15] M.A. Lago, A.D.A. Maidment, P.R. Bakic, "Modelling of mammographic compression of anthropomorphic software breast phantom using FEBio," *Proc. Int'l Symposium on Computer Methods in Biomechanics and Biomedical Engineering* (CMBBE) Salt Lake City, UT, 2013.
[16] J. Kuo, P. Ringer, S.G. Fallows, S. Ng, P.R. Bakic, A.D.A. Maidment, "Dynamic reconstruction and rendering of 3D tomosynthesis images" *Proc. of SPIE,* Medical Imaging 2011.
[17] J.D. Sain, and H.H. Barrett, "Performance Evaluation of a Modular Gamma Camera Using a Detectability Index," *J Nucl Med*, vol. 44, no. 1, pp. 58-66, Jan 2003.